\documentclass[11pt]{article}
\usepackage{multirow}
\usepackage{soul}
\usepackage{amsmath}
\usepackage{float}

\usepackage[final]{acl}

\usepackage{times}
\usepackage{latexsym}

\usepackage[T1]{fontenc}

\usepackage[utf8]{inputenc}

\usepackage{microtype}

\usepackage{amsmath, amsthm, amssymb, graphicx}

\usepackage{inconsolata}

\usepackage{graphicx}

\title{Turn-level Multiscale Density Ratio Estimation for LLM Agents}

\author{Zishuo Zhao$^{1}$ , \ Kai Chen$^{1}$ , \ Ao Li$^{1}$  , \ Yuan Liu$^{1,*}$ 
\\\textsuperscript{\rm 1}Alibaba Group \\
  \texttt{\{zhaozishuo.zzs,muzi.la,ck489728,xuanjing.ly\}@alibaba-inc.com} }

\begin{document}

\maketitle
\renewcommand*{\thefootnote}{\fnsymbol{footnote}}
\footnotetext[1]{Corresponding author.} 
\begin{abstract}
With the rapid development of Large language model (LLM), agent systems enhanced by LLMs show huge potential in being able to deal with complex tasks, especially involving multi-step thinking or interaction with tools. For applying LLM techniques with a well-designed agent paradigm, post-training of LLM in multiple agent scenarios is necessary to achieve better performance. Among the variable post-training techniques, alignment methods such as PPO, DPO, DIL, and GRPO become popular because many papers show a significant positive impact on the model's performance by punishing negative samples while keeping acceptable training complexity. However, most alignment methods address simple single-turn tasks, and there remains room for improvement for complex multi-turn tasks. We propose Turn-level Multiscale Density Ratio Estimation (\textbf{tlm-DRE}), which assigns different weights on corresponding turns and proposes asymmetric token-level training based on the positive-negative space gaps across multiple turns of tasks. The results of the experiment on a wide range of agent benchmarks show that the proposed method performs competitively compared to traditional alignment methods. The proposed training method enables LLMs to perform robustly in multi-turn reasoning tasks with both in-domain and out-of-domain conditions.
\end{abstract}

\section{Introduction}

Enhancing agents' capabilities to tackle diverse complex tasks, which often involves interacting with a sophisticated environment equipped with a bunch of tools, has attracted considerable attention. For example, such tasks include complex social dialogue \citep{wang2023rolellm,park2023generative}, scientific experiment \citep{wang2022scienceworld}, embodied housework \citep{shridhar2020alfworld, li2024behavior}, multi-hop question answering \citep{yang2018hotpotqa, ho-etal-2020-constructing}, etc.

To accomplish these tasks, LLM-based agents must interact with the environment step by step, decomposing the final goal into sub-goals, and then plan next action based on feedback from the environment. Research on LLM-based agents initially focused on directly generating trajectories using large language models. Most studies employ prompt engineering to enhance the trajectory generation capabilities of large language models, such as CoT \citep{wei2022chain}, ReAct \citep{yao2022react}, and Reflexion \citep{shinn2023reflexion}. Subsequent research focused on trajectory tuning to further enhance agent planning capabilities \citep{chen2023fireact,yin2023agent}.

Meanwhile, reinforcement learning of Large Language Models \citep{achiam2023gpt, touvron2023llama, bai2023qwen} becomes one of the most important post-training approaches \citep{kumar2022llm} to tune LLMs more applicable to overcome shortcomings, such as hallucinations and logical consistency. Several efficient alignment training methods have been proposed in recent research discourse \citep{ouyang2022training, rafailov2023direct, shao2024deepseekmath}. Such post-training paradigms have also been applied to LLM-based agent tasks recently, trying to overcome the drawbacks of simply utilizing the zero-shot LLMs, which neglect agent training.

More specifically, LLM-based agent tasks typically employ the heuristic model(e.g., GPT-4) to generate a group of expert trajectories with a certain CoT form and a set of sampling strategies as a filter. Further supervised fine-tuning (SFT) is then launched to enhance the model's reasoning and planning adaptation to certain domains. Driven by the SFT-trained reference policy, more trajectories are sampled \citep{song2024trial,shi2024direct,xiong2024watch,kong2025sdpo} and rewarded step by step to evaluate the capacity gaps of the reference model through feedback from the simulated environment. Subsequently, an alignment approach such as RHLF \citep{ouyang2022training}, DPO \citep{rafailov2023direct},  DRE \citep{xiao2025on}, GRPO \citep{shao2024deepseekmath} will be applied as a key role in those tasks to calibrate the sampling distribution by penalizing low-quality trajectories while preserving the original probability mass over high-quality samples.

However, existing methods of alignment on agent tasks still exhibit discrepancies for future development. For example, the aforementioned RL approaches perform alignment directly at the trajectory level while being lack of attention to turn-level details, resulting in suboptimal overall alignment performance. Furthermore, even though a new LLM alignment paradigm has been put forward that views the process as a typical imitation learning under the framework of density ratio estimation, studies on the performance of this "imitation type" of alignment in agent-based tasks remain relatively scarce. Aimed to address these challenges, we propose Turn-level Multiscale Density Ratio Estimation (\textbf{tlm-DRE}) for LLM Agents, which applies imitation learning on agent-based tasks with turn-level alignment efficiency.

In particular, we assign lower weights to the turns of the samples with higher policy confidence, since there is no need to further align between chosen and rejected sample pairs. On the other hand, we assign higher weights to the turns with lower policy confidence since those turns are more crucial to be calibrated in the language space. Based on the turn weights derived from the reference policy, we design a novel definition of density ratio estimation under varying turn-level scales. Afterwards, we launch the alignment over extensive agent-based tasks in the framework of imitation learning.  

Our contributions are as follows: (1) We propose a novel \textbf{tlm-DRE} method that systematically applies imitation learning to agent-based tasks in a turn-level multiscale probability space. (2) We conduct extensive experiments on several agent-based tasks \citep{shridhar2020alfworld, wang2022scienceworld, yang2018hotpotqa, zhang2026the} to show that our method surpasses current methods with stable and robust performance. (3) We present a comprehensive analysis to support the efficacy of our method from various points of view.

\section{Related Work}

\paragraph{LLM application in agent-based tasks}

The development of LLM has inspired intelligent agents to design complex tasks involved with a dynamic environment and a complicated real-world toolbox. The main motivation to utilize LLM in those tasks is that the agents need strong abilities of reasoning(both externally-oriented and reflectional) and planning. CoT \citep{wei2022chain} enables LLM to articulate its own thought processes, enhancing its reasoning capabilities and laying the groundwork for subsequent agent reasoning frameworks. ReAct \citep{yao2022react} integrates feedback from the environment into its reasoning process, allowing the model to think when taking actions, interacting with the environment, and adjusting subsequent actions. Reflexion \citep{shinn2023reflexion} builds on ReAct by incorporating LLM self-reflection, allowing the model to autonomously correct its previous erroneous actions. There are even more complicated multi-agent designs \citep{park2023generative} that simulate believable human behavior in the simulated sandbox.

\paragraph{Alignment process: Reinforcement Learning}

\begin{figure*}[!htb]
    \centering
    \center{\includegraphics[width=2.10\columnwidth]{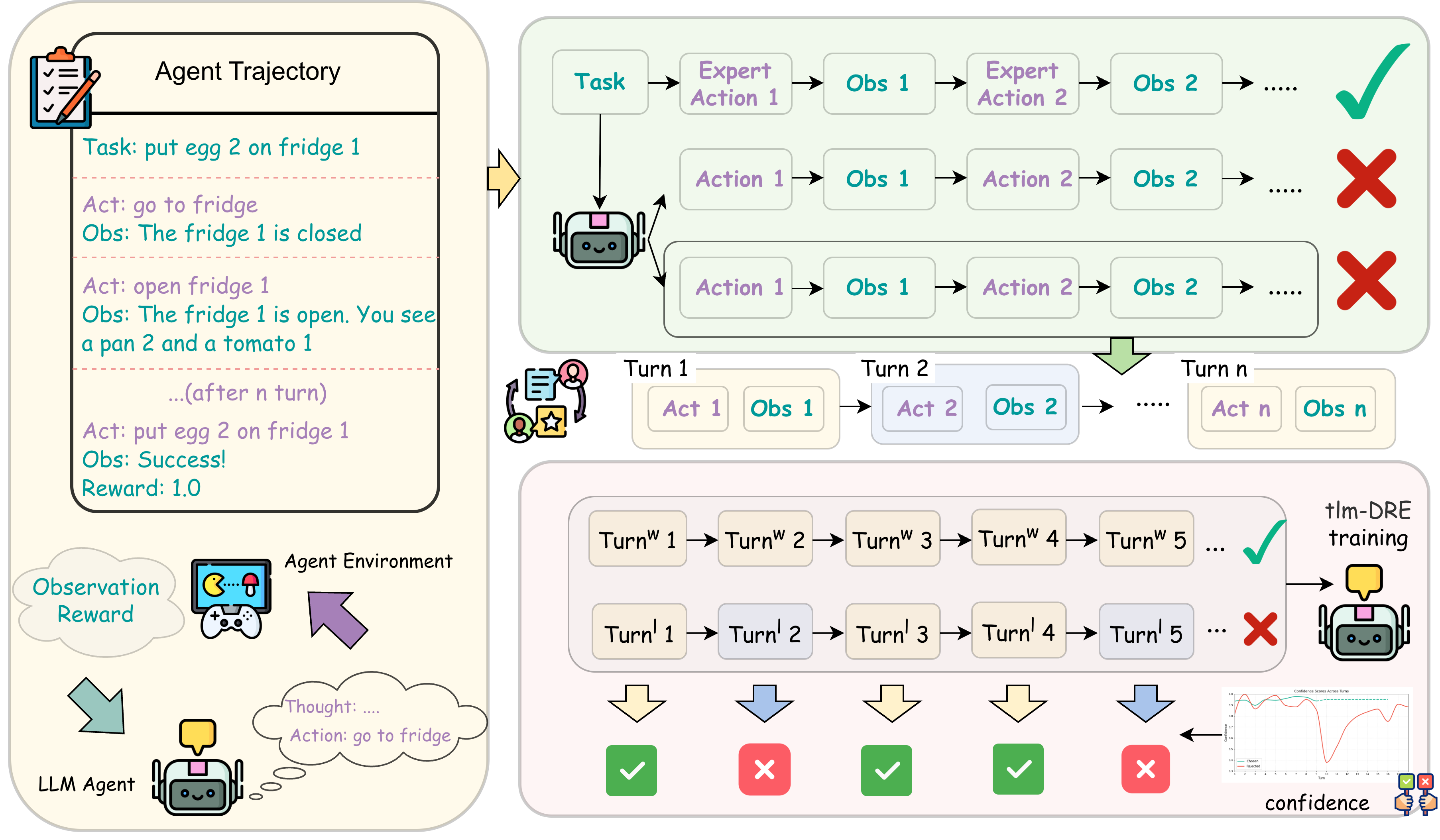}}
    
    \caption{The overall architecture of \textbf{tlm-DRE} in a single iteration. First, the agent initialized with the SFT policy samples trajectory paths and collects failure trajectories. Then, within these failure trajectories, we identify low-confidence turns—i.e., turns where the model exhibits low self-confidence. Finally, we train the agent using Multiscale Density Ratio Estimation, which explicitly upweights these identified failure turns during optimization.}
    \label{fig:paper1}
\end{figure*}

Many researchers focus on using reinforcement learning to further enhance agent performance. ETO \cite{song2024trial} introduces Direct Preference Optimization for post alignment. They use the SFT model as a reference policy to generate bad cases, which are then paired with expert trajectories to form sample pairs. DMPO \cite{shi2024direct} has noted that training directly with DPO leads to mismatching trajectory step lengths between positive and negative samples, and length regularization has been proposed. In contrast, SDPO \cite{kong2025sdpo} directly utilizes GPT to capture positive and negative samples of equal step lengths, avoiding the issue of mismatched step length. 

Some research approaches agent training using online RL, such as GRPO \cite{shao2024deepseekmath}, GSPO \cite{zheng2025group}. Such kind of methods are group-based RL approaches that are critic-free, making training simple and stable. GiGPO \cite{feng2025group} groups objects that share identical environmental interaction states, employing this hierarchical Group-in-Group structure to train GRPO. Other methods \citep{wei2025reinforcing, DBLP:journals/corr/abs-2505-20023} also try to apply strategies such as turn-weight or reflection on agentic tasks. These methods have also shown to be highly effective in post-alignment.

\paragraph{Alignment process: Imitation Learning}

Some research focuses on applying imitation learning(IR) in the artificial intelligence domain. The key idea of IR is to directly extract knowledge from demonstrations by human experts or artificially created agents. Inverse Reinforcement Learning has been proposed to recover the reward function of empirical resources from the uncertain environment \citep{russell1998learning, ng2000algorithms, sun2024inverse}. Behavioral cloning \citep{pomerleau1991efficient, ross2011reduction} efficiently bridges the domain gap by supervised fine-tuning on expert trajectories. The adversarial Imitation Learning method, such as GAIL \citep{ho2016generative}, AIRL \citep{fu2018learning}, leverages adversarial training with online interaction with the environment. 

Density Ratio Estimation \citep{sugiyama2012density, amari2010information} can be viewed as a typical way to mimic the behavior of experts under a certain distribution space. DRE can be measured by optimizing the discrepancy between the learnable policy and the ideal distribution equipped with the Bregman divergence. Some methods have already tried to transplant this idea into the post training of LLM based tasks, such as DIL \citep{xiao2025on}, GSIL \citep{xiao2024leverage}. Furthermore, DIL \citep{xiao2025on} also shows that traditional reinforcement learning methods, such as RLHF and DPO, are actually a form of imitation learning.

\section{Notations and Preliminaries}

\subsection{Task Formulation}
Consider one agent which is designed to deal with a certain bunch of complex tasks that interact with the real world or simulation environment. Let $o_i \in \mathcal{O}$ represent the observation of the agent for the $i$-th turn of interaction with the environment. When $i=1$, $o_i$ represents the initial description of a given task and the initial status of the agent. Based on the agent's observation from the environment, as well as the previous context or trajectory $c_t = (o_1, a_1, ..., o_{t-1}, a_{t-1}, o_t)$, a type of action from the action space $\mathcal{A}$ will be chosen after a reasoning process. To avoid redundancy, we might as well view the thinking process and the action that the agent takes as a whole, named $a_t \in \mathcal{A}$ for $t$ turns, which follows a parameterized policy $\pi_{\theta}(a_t|c_t)$. 

For simplicity, the potential stochastic effect of how the action is being executed is neglected. Furthermore, if the policy $\pi$ is driven by a certain large language model, then the action space $\mathcal{A}$ is equivalent to a language space $\mathcal{L}$. In detail, the whole trajectory $c_T$ can be viewed as $(\boldsymbol{x}, \boldsymbol{y_T})$, where $\boldsymbol{x} = [x_1, x_2, ...]$ belongs to $\mathcal{L}$ while $x_i$ is the language token. Similarly, $\boldsymbol{y_T} = [\boldsymbol{y_{a_{1}}}, \boldsymbol{y_{o_{2}}}, \boldsymbol{y_{a_{2}}}, ..]$, $\boldsymbol{y_{o_{i}}}$, $\boldsymbol{y_{a_{i}}}$ also belong to the language space $\mathcal{L}$. 

Denote $\pi_{ref}$ as the reference policy. Based on the reference policy and the given input language sequences $\boldsymbol{x}$, trajectories $\boldsymbol{y}$ will be sampled following a sample strategy. Denote $\boldsymbol{y_{w}}$ as the preferred trajectory which is marginally superior to another trajectory $\boldsymbol{y_{l}}$, based on the final behavior and the evaluation of the whole process. Suppose that the ideal optimized policy for the entire policy space is $\pi_{c}$. The object is to find the optimized projected policy $\pi_{\theta}$ for a parameterized policy space.

\subsection{Direct Preference Optimization}

For standard reinforcement learning from human feedback(RLHF), a reward model is trained to evaluate whether the policy behaves well enough, such as PPO.  DPO \cite{rafailov2023direct} proposes a way to directly use $\pi_\theta / \pi_{ref}$ as the reward function $r(\boldsymbol{y} | \boldsymbol{x})$ and interpret the alignment process as a task to optimize the margin between the pair-wise samples.

\paragraph{Bradley-Terry Model}

Bradley-Terry model is applied to measure the partial order relation of a sample pair $(\boldsymbol{y_w}, \boldsymbol{y_l})$ stipulating the human preference distribution $p^{*}$ as: 
\begin{equation} \label{bt_model}
p^{*}(\boldsymbol{y_w} \succ \boldsymbol{y_l}) = \sigma(r(\boldsymbol{x}, \boldsymbol{y_w}) - r(\boldsymbol{x}, \boldsymbol{y_l})),
\end{equation}
where $\sigma$ is the sigmoid function. 

Assuming a static data set of pair-wise samples $\mathcal{D} = \{\boldsymbol{x}^i, \boldsymbol{y}^i_w, \boldsymbol{y}^i_l\}^N_{i=1}$ exists, then the DPO method will try to find a reward model $r_\theta$ in a parametrized model space that minimizes the following negative log-likelihood loss:
\begin{equation} \label{dpo_loss}
\begin{aligned} 
\mathcal{L}_{DPO} &= -\mathbb{E}_{(\boldsymbol{x}, \boldsymbol{y}_w, \boldsymbol{y}_l \sim \mathcal{D})} [log\sigma(r_{\theta}(\boldsymbol{y}_w) - \\
& r_{\theta}(\boldsymbol{y}_l))]  + \beta \mathbb{D}_{kl}[\pi_\theta \Vert \pi_{ref}].
\end{aligned} 
\end{equation}

\subsection{Density Ratio Estimation}

\paragraph{Bregman Divergence}
Denote the density ratio as $r_{\theta}$ as $\frac{\pi_{\theta}}{\pi_{ref}}$, the true density ratio is $r^{*}$ as $\frac{\pi_{c}}{\pi_{ref}}$. To employ the Bregman divergence(BR) for estimating the density ratio according to \cite{sugiyama2012density}, let $f$ be a differentiable and strictly convex function in the density-ratio model space. the discrepancy from the true density ratio $r^{*}$ to the density-ratio model $r_{\theta}$ can be measured as:

\begin{equation} \label{eq_breg_dre}
\begin{aligned} 
BR_{f}^{\prime}(r^{*} & \Vert r_{\theta})  =  ( f(r^{*}) - f(r_{\theta}) \\  & - \partial f(r_{\theta}) (r^{*} - r_{\theta} ) ).
\end{aligned}
\end{equation}

Fig.~\ref{fig:bregman_divergence} in Appendix~\ref{appendix_kernal} illustrates how the current point $r_\theta$ can slide towards the target point $r^{*}$ iteratively by optimizing the Bregman divergence. Several kernel functions $f$ for Bregman divergence can be found in Appendix~\ref{appendix_kernal}.

\paragraph{Directly Imitation Learning}

Under the assumption that we have a datasets $\mathcal{D}$ as defined above, the DIL method \citep{xiao2025on} tries to imitate the optimal density ratio by minimizing the discrepancy under the Bregman divergence in a parameterized constriction:
\begin{equation} \label{eq_breg_dre_discrete}
\begin{aligned} 
\mathop{\mathrm{min}}\limits_{\theta} D_h(r^* \Vert r_\theta) & = \sum_{\boldsymbol{y}}\pi_{ref}( f(r^{*}) - f(r_{\theta}) \\  & - \partial f(r_{\theta}) (r^{*} - r_{\theta} ) ).
\end{aligned}
\end{equation}
Removing the irrelevant constant $\theta$ and under the assumption that $\boldsymbol{y}_l \sim \pi_{ref}(\boldsymbol{y}_l \Vert \boldsymbol{x})$, we then get the loss function of the DIL method:
\begin{equation} \label{dil_loss}
\begin{aligned} 
\mathcal{L}_{DIL} &= - \mathbb{E}_{(\boldsymbol{x}, \boldsymbol{y}_w, \boldsymbol{y}_l \sim \mathcal{D})} \{ \partial f(r_{\theta}(\boldsymbol{y}_w)) - \\ & [\partial f(r_\theta(\boldsymbol{y}_l))r_\theta(\boldsymbol{y}_l) - f(r_\theta(\boldsymbol{y}_l))] \}.
\end{aligned} 
\end{equation}

\section{Methodology}

Due to varying degrees of training exposure of tokens in the embedding space within the reference model, one daunting challenge is how to train the policy over the candidate tokens more efficiently. This problem warrants more attention for the tasks that need multi-step thinking and reasoning. \cite{ICLR2025_7fb9f390} introduces a way to allocate token weights to each token according to a certain importance measurement for the DPO. However, this idea cannot be trivially generalized to turn-level optimization for the DRE loss due to the fact that most of the Bregman divergence functions show a high degree of nonlinearity and asymmetry. We will introduce a new density ratio function that will be used to solve this problem. Our method framework is illustrated in Figure~\ref{fig:paper1}.

\subsection{Multiscale Density Ratio Representation}

Since the density ratio is designed to estimate how close an approximation to the optimal solution can be achieved over the output language space, we can only focus on the results of linguistic action $\boldsymbol{y_{a_{i}}}$ under the assumption that the observation results are determined by omitting the stochastic effect from the simulation environment. More specifically, the density ratio can be expressed as follows:
\begin{equation} \label{dre_expression}
\begin{aligned} 
r(& \boldsymbol{y_t} \Vert \boldsymbol{x})  = \frac{\pi(\boldsymbol{y_t} \Vert \boldsymbol{x})}{\pi_{ref}(\boldsymbol{y_t} \Vert \boldsymbol{x})} \\ & = \frac{\pi(\boldsymbol{y_{a_t}} \Vert \boldsymbol{x}, \boldsymbol{c_{t-1}} ) ... \pi(\boldsymbol{y_{a_1}} \Vert \boldsymbol{x})}{\pi_{ref}(\boldsymbol{y_{a_t}} \Vert \boldsymbol{x}, \boldsymbol{c_{t-1}} ) ... \pi_{ref}(\boldsymbol{y_{a_1}} \Vert \boldsymbol{x})},
\end{aligned}
\end{equation}
where $\boldsymbol{c}_{t-1} = [\boldsymbol{y_{a_1}}, ...,\boldsymbol{y_{o_{t-1}}}]$.

The main drawback of the traditional expression of the DRE is the inefficiency of training the underfitting tokens, whereas well-trained tokens will possibly suffer the risk of overfitting. Meanwhile, as demonstrated in \cite{ICLR2025_7fb9f390}, the importance of each token in latent embedding spaces is not uniformly distributed, implying heterogeneous training demands. In the agent-related tasks with multi-step reasoning, the importance weighting with non-uniformity becomes more pronounced from the turn-level perspective.

Suppose that a turn-level weighted function: 
\begin{equation}
\begin{aligned}
\omega_t = \omega(\boldsymbol{y_{a_t}} \Vert \boldsymbol{x}, \boldsymbol{c_{t-1}} ),
\end{aligned}
\end{equation}
measures how important and how well-trained each turn in the agentic reasoning process is. The density ratio then has to be measured as:
\begin{equation} \label{dre_expression_weighted}
\begin{aligned} 
r(& \boldsymbol{y_t} \Vert \boldsymbol{x})  = \frac{\hat{\pi}(\boldsymbol{y_t} \Vert \boldsymbol{x})}{\hat{\pi}_{ref}(\boldsymbol{y_t} \Vert \boldsymbol{x})} \\ & = \frac{\pi^{\omega_t}(\boldsymbol{y_{a_t}} \Vert \boldsymbol{x}, \boldsymbol{c}_{t-1} ) ... \pi^{\omega_1}(\boldsymbol{y_{a_1}} \Vert \boldsymbol{x})}{\pi^{\omega_t}_{ref}(\boldsymbol{y_{a_t}} \Vert \boldsymbol{x}, \boldsymbol{c}_{t-1} ) ... \pi^{\omega_1}_{ref}(\boldsymbol{y_{a_1}} \Vert \boldsymbol{x})}.
\end{aligned}
\end{equation}

\subsection{Turn-level Weighted Measurement}

We define the adequacy estimate of the $i$-th turn output $\boldsymbol{y_i}$ as $p_i$:
\begin{equation} \label{Adequacy_Estimation}
\begin{aligned} 
  \mathop{\mathrm{log}}  & p_i  = \frac{1}{|\boldsymbol{y_i}|} \mathop{\mathrm{log}}\pi_{ref}(\boldsymbol{y}_{a_i} \Vert \boldsymbol{x}, \boldsymbol{c}_{i-1})\\ & = \frac{1}{|\boldsymbol{y_i}|} \sum^{|\boldsymbol{y}_i|}_{j=1}\mathop{\mathrm{log}\pi_{ref}(y_{i,j}|\boldsymbol{x}, \boldsymbol{c}_{i-1}, y_{i,<j})},
\end{aligned}
\end{equation}
where $|\boldsymbol{y}_i|$ is the total number of tokens $\boldsymbol{y}_i$.

In some previous searches \cite{nagumo2024density}, they view the statistical possibility of the output as an outlier measurement. However, we argue that, under the assumption that previous training is inefficient and imbalanced, the adequacy estimation actually shows more weight on how well trained the $i$-th turn of the trajectory is, based on the reference policy , rather than the outliers measurement. Hence, a natural way is to make sure that there is a negative correlation between the adequacy estimation and the turn-weights applied on the following alignment.

We define the turn-level weights $\omega_i$ for the i-th turn of the trajectory as
\begin{equation} \label{turn_weight}
\begin{aligned} 
\omega_i = \left\{
\begin{array}{l}
    \omega_{L} \quad p_i > p_{pivot},\\
    \omega_{U} \quad p_i \leq p_{pivot}.
\end{array}
\right.
\end{aligned}
\end{equation}

\subsection{Turn-level Weighted DRE Optimization}
Applying the turn-level weight version of the density ratio as defined in Equation (\ref{dre_expression_weighted}) to the DRE optimization, Equation (\ref{eq_breg_dre_discrete}) becomes:
\begin{equation} \label{eq_breg_dre_discrete_weighted}
\begin{aligned} 
\mathop{\mathrm{min}}\limits_{\theta} D_{\omega, f}&(\hat{r}^* \Vert \hat{r}_\theta)  = \sum_{\boldsymbol{y}}\pi_{ref}( f(\hat{r}^{*}) \\  & - f(\hat{r}_{\theta}) - \partial f(\hat{r}_{\theta}) (\hat{r}^{*} - \hat{r}_{\theta} ) ).
\end{aligned}
\end{equation}
Subtracting the constant $\sum_{\boldsymbol{y}}\pi_{ref}f(\hat{r}^{*})$, we obtain:
\begin{equation} \label{eq_breg_dre_discrete_weighted_deduce_1}
\begin{aligned} 
\mathop{\mathrm{min}}\limits_{\theta} D_{\omega, f}&(\hat{r}^* \Vert \hat{r}_\theta)  =  - \sum_{\boldsymbol{y}}\pi_{ref}\partial f(\hat{r}_{\theta})\hat{r}^* \\ & + \sum_{\boldsymbol{y}}\pi_{ref} (\partial f(\hat{r}_{\theta})\hat{r}_{\theta} -  f(\hat{r}_{\theta})).
\end{aligned}
\end{equation}
The last term above $\pi_{ref}\partial f(\hat{r}_{\theta})\hat{r}^*$ can be expanded as:
\begin{equation} \label{eq_breg_dre_discrete_weighted_deduce_2}
\begin{aligned} 
& \qquad\pi_{ref}\partial f(\hat{r}_{\theta})\hat{r}^* \\ & = \pi_{ref}\partial f(\hat{r}_{\theta})
\prod\limits_{i=1}^{t}\frac{\pi^{\omega_i}_{c}(\boldsymbol{y_{a_i}} \Vert \boldsymbol{x}, \boldsymbol{c}_{i-1} )}{\pi^{\omega_i}_{ref}(\boldsymbol{y_{a_i}} \Vert \boldsymbol{x}, \boldsymbol{c}_{i-1} )}
\\ & = \partial f(\hat{r}_{\theta})\prod\limits_{i=1}^{t}\frac{\pi^{\omega_i}_{c}(\boldsymbol{y_{a_i}} \Vert \boldsymbol{x}, \boldsymbol{c}_{i-1} )}{\pi^{\omega_i-1}_{ref}(\boldsymbol{y_{a_i}} \Vert \boldsymbol{x}, \boldsymbol{c}_{i-1} )}.
\end{aligned}
\end{equation}
Due to the fact that the possibility of the un-chosen trajectory by policy $\pi_{c}$ is vanishingly small and $\omega \in [\omega_L, \omega_U]$, Substituting (\ref{eq_breg_dre_discrete_weighted_deduce_2})into (\ref{eq_breg_dre_discrete_weighted_deduce_1}) the equation yields:
\begin{equation} \label{eq_breg_dre_discrete_weighted_deduce_3}
\begin{aligned} 
& \qquad \mathop{\mathrm{min}}\limits_{\theta} D_{\omega, f}(\hat{r}^* \Vert \hat{r}_\theta) \\ & = \sum_{\boldsymbol{y}}\pi_{ref} (\partial f(\hat{r}_{\theta})\hat{r}_{\theta} -  f(\hat{r}_{\theta})) \\ & - \sum_{\boldsymbol{y}_w}\pi_{c}\partial f(\hat{r}_{\theta})\prod\limits_{i=1}^{t}\frac{\pi^{\omega_i-1}_{c}(\boldsymbol{y_{a_i}} \Vert \boldsymbol{x}, \boldsymbol{c}_{i-1} )}{\pi^{\omega_i-1}_{ref}(\boldsymbol{y_{a_i}} \Vert \boldsymbol{x}, \boldsymbol{c}_{i-1} )},
\end{aligned}
\end{equation}
under the same assumption that $\boldsymbol{y}_l \sim \pi_{ref}(\boldsymbol{y}_l \Vert \boldsymbol{x})$ the loss function of (\textbf{tlm-DRE}) is:
\begin{equation} \label{tm_dre_loss_deduce}
\begin{aligned} 
& \qquad \mathcal{L}_{tlm-DRE} \\ & = \int_{\boldsymbol{y}_l} [\partial f(\hat{r}_\theta(\boldsymbol{y}_l))\hat{r}_\theta(\boldsymbol{y}_l) - f(\hat{r}_\theta(\boldsymbol{y}_l))] \\ & - \int_{\boldsymbol{y}_w} \partial f(\hat{r}_{\theta})\prod\limits_{i=1}^{t}\frac{\pi^{\omega_i-1}_{c}(\boldsymbol{y_{a_i}} \Vert \boldsymbol{x}, \boldsymbol{c}_{i-1} )}{\pi^{\omega_i-1}_{ref}(\boldsymbol{y_{a_i}} \Vert \boldsymbol{x}, \boldsymbol{c}_{i-1} )}.
\end{aligned} 
\end{equation}
The distribution shows a significant concentration on the candidate tokens of the chosen trajectory in the language space. Since $\pi_{c}$ is the ideal policy to be approximated from the parameterized policy's manifold, we might as well assume that $\pi_{c}(\boldsymbol{y_{a_i}}\Vert \boldsymbol{x}, \boldsymbol{c}_{i-1} ) \sim 1$ uniformly, leading to the final \textbf{tlm-DRE} loss for training: 
\begin{equation} \label{tm_dre_loss}
\begin{aligned} 
& \qquad \mathcal{L}_{tlm-DRE} \\ & = \mathbb{E}_{(\boldsymbol{x}, \boldsymbol{y}_w, \boldsymbol{y}_l \sim \mathcal{D})} \{ [\partial f(\hat{r}_\theta(\boldsymbol{y}_l))\hat{r}_\theta(\boldsymbol{y}_l) - f(\hat{r}_\theta(\boldsymbol{y}_l))]
\\ & - \partial f(\hat{r}_{\theta}(\boldsymbol{y}_w))\prod\limits_{i=1}^{t}\frac{\pi^{\omega_i-1}_{c}(\boldsymbol{y_{w, a_i}} \Vert \boldsymbol{x}, \boldsymbol{c}_{w, i-1} )}{\pi^{\omega_i-1}_{ref}(\boldsymbol{y_{w, a_i}} \Vert \boldsymbol{x}, \boldsymbol{c}_{w, i-1} )} \} 
\\ & \approx \mathbb{E}_{(\boldsymbol{x}, \boldsymbol{y}_w, \boldsymbol{y}_l \sim \mathcal{D})} \{ [\partial f(\hat{r}_\theta(\boldsymbol{y}_l))\hat{r}_\theta(\boldsymbol{y}_l) - f(\hat{r}_\theta(\boldsymbol{y}_l))]
\\ & - \partial f(\hat{r}_{\theta}(\boldsymbol{y}_w))\prod\limits_{i=1}^{t}\frac{1}{\pi^{\omega_i-1}_{ref}(\boldsymbol{y_{w, a_i}} \Vert \boldsymbol{x}, \boldsymbol{c}_{w, i-1} )} \}.
\end{aligned} 
\end{equation}

In this paper, we use UKL \cite{nguyen2010estimating} defined in Appendix~\ref{appendix_kernal} as the kernel function for Bregman divergence. Then equation (\ref{tm_dre_loss}) will be as:
\begin{equation} \label{tm_dre_loss_ukl}
\begin{aligned} 
& \qquad \mathcal{L}_{tlm-DRE} 
\\ & = \mathbb{E}_{(\boldsymbol{x}, \boldsymbol{y}_w, \boldsymbol{y}_l \sim \mathcal{D})} \{ \prod_{i}\frac{\pi_\theta^{\omega_{l, i}}(\boldsymbol{y}_{l, i})}{\pi_{ref}^{\omega_{l, i}}(\boldsymbol{y}_{l, i})} -
\\ & \sum_{i} \frac{\omega_{w, i}}{\prod_{i=1}^{t}\pi^{\omega_i-1}_{ref}(\boldsymbol{y_{w, i}})} \mathop{\mathrm{log}}\frac{\pi_\theta(\boldsymbol{y}_{w, i})}{\pi_{ref}(\boldsymbol{y}_{w, i})}  \}.
\end{aligned} 
\end{equation}

\section{Experiments}
In this section, we demonstrate the performance of \textbf{tlm-DRE} across various agent based tasks, provide detailed experimental procedures, and introduce other related baselines.

\subsection{Experimental Settings}

\begin{table*}[t]
\centering
\setlength{\tabcolsep}{7pt}
\begin{tabular}{clcccc}
\hline
\multirow{2}{*}{Paradigm}
& \multicolumn{1}{c}{\multirow{2}{*}{Models}}
& \multicolumn{2}{c}{ALFWorld}
& \multicolumn{2}{c}{ScienceWorld} \\
& & Seen & Unseen & Seen & Unseen \\
\hline

\multirow{3}{*}{Prompt-based}
& GPT-3.5 \cite{ouyang2022training}
& 7.9 & 10.5 & 16.5 & 13.0 \\
& GPT-4 \cite{achiam2023gpt}
& 42.9 & 38.1 & 64.8 & 64.4 \\
& Qwen2.5-7B \cite{team2024qwen2}
& 25.1 & 28.4 & 26.8 & 25.2 \\
\hline

\multicolumn{6}{l}{\textit{Llama-2-7B-Chat}} \\
\multirow{6}{*}{SFT \& RL}
& SFT \cite{chen2023fireact}
& 60.0 & 67.2 & 56.8 & 56.0 \\
& PPO \cite{trung2024reft}
& 22.1 & 29.1 & 59.4 & 51.7 \\
& RFT \cite{zhang2023cumulative}
& 62.9 & 66.4 & 71.6 & 54.3 \\
& ETO \cite{song2024trial}
& 68.6 & 72.4 & 68.5 & 61.1 \\
& DMPO \cite{shi2024direct}
& 43.3 & 55.0 & \textbf{72.4} & \textbf{61.7} \\
& tlm-DRE (ours)
& \textbf{70.0} $\pm {\scriptstyle 0.71}$
& \textbf{72.6} $\pm {\scriptstyle 0.49}$
& 71.4 $\pm {\scriptstyle 1.27}$
& 61.2 $\pm {\scriptstyle 0.70}$ \\
\hline

\multicolumn{6}{l}{\textit{Qwen2.5-7B-Instruct}} \\
\multirow{6}{*}{RL training}
& SFT \cite{chen2023fireact}
& 70.7 & 83.6 & 71.8 & 61.8 \\
& ETO \cite{song2024trial}
& 75.0 & 86.6 & 69.1 & 62.8 \\
& DIL \cite{xiao2025on}
& 73.6 & 88.8 & 71.9 & 63.7 \\
& tlm-DRE w/o DRE
& 74.8 $\pm {\scriptstyle 0.47}$
& 87.3 $\pm {\scriptstyle 0.75}$
& 72.3 $\pm {\scriptstyle 0.59}$
& 63.4 $\pm {\scriptstyle 1.53}$ \\
& tlm-DRE w/o tlm
& 75.0 $\pm {\scriptstyle 1.43}$
& 89.3 $\pm {\scriptstyle 0.99}$
& 72.3 $\pm {\scriptstyle 0.99}$
& 63.9 $\pm {\scriptstyle 0.81}$ \\
& tlm-DRE (ours)
& \textbf{75.5} $\pm {\scriptstyle 0.47}$
& \textbf{90.1} $\pm {\scriptstyle 0.50}$
& \textbf{72.7} $\pm {\scriptstyle 0.75}$
& \textbf{64.5} $\pm {\scriptstyle 0.54}$ \\
\hline
\end{tabular}

\caption{
Performance of different methods on ALFWorld and ScienceWorld,
reported as average reward.
``Seen'' denotes the held-out test set containing task types observed
during training, while ``Unseen'' refers to test tasks with critical
unseen variations (e.g., novel objects or goals).
``tlm-DRE w/o DRE'' denotes linearly turn-weighted DPO training.
For a fair comparison, all methods use the same base models:
Llama-2-7B and Qwen2.5-7B.
All experimental results are averaged over five independent runs
with different random seeds.
}
\label{table1}
\end{table*}

\paragraph{Datasets} We perform experiments across a diverse set of environments to evaluate the capabilities of our agent. Specifically, we use SciWorld \cite{wang2022scienceworld} under the Apache-2.0 license for grounded scientific experimentation in simulated laboratories, ALFWorld under the MIT License \cite{shridhar2020alfworld} for embodied household tasks requiring object manipulation and planning in 3D environments. In addition, we also include HotpotQA \cite{yang2018hotpotqa} under the CC BY-SA 4.0 license as the benchmark for multi-hop reasoning in open-domain settings, a task that requires the agent to retrieve and integrate information from multiple sources to answer complex questions. SciWorld provides dense final rewards on a continuous scale from 0 to 1, whereas ALFWorld offers only sparse binary rewards that indicate whether the task was completed or not. For multi-hop QA tasks, we measure reasoning performance using EM and F1 scores against ground-truth answers.

\paragraph{Baselines} We compare our approach with a series of benchmarks: (1) Zero-shot using LLMs such as GPT-4o, Qwen2.5-7B, applying the ReAct prompting paradigm, which represents the
state-of-the-art zero-shot capabilities of LLMs.(2) SFT (Supervised Fine-Tuning) conducts behavioral cloning on expert trajectories. (3) PPO (Proximal Policy Optimization) as an actor-critic reinforcement learning algorithm to directly optimize the initialized SFT policy. (4) ETO \cite{song2024trial} uses successful and failure trajectories as sample pairs for DPO training. (5) DMPO \cite{shi2024direct} adds length regularization to ETO to eliminate noise caused by failure trajectories with excessive steps. (6) DIL \cite{xiao2025on} applies the DRE to the alignment learning stage from the perspective of imitation learning. 

All experiments were conducted on 8x Nvidia A100 GPUs, each with 80GB of memory, implemented using PyTorch in Python. Details of the experiment setup and how we set the hyperparameters can be found in the Appendix~\ref{appendix_setup}. 

\begin{table}
\centering
\setlength{\tabcolsep}{10pt}
\begin{tabular}{ccc}
\hline
 & \multicolumn{2}{c}{HotpotQA} \\
\multirow{-2}{*}{Method} & EM & F1 \\ \hline
SFT \cite{chen2023fireact} & 27.80 & 36.45 \\
CoH \cite{liu2023chain} & 28.60 & 39.53 \\
PPO \cite{trung2024reft} & 28.20 & 36.47 \\
DPO \cite{song2024trial} & 26.40 & 34.83 \\
NAT \cite{wang2025nat} & 29.60 & 42.50 \\ \hline
tlm-DRE (ours) & \textbf{33.8} & \textbf{43.74} \\ \hline
\end{tabular}
\caption{Overall results on open-domain question answering tasks. We measure the performance using Exact Match and F1 score. For fair comparison, all methods use the same base models: LLaMA2-7B}
\label{tabel3}
\end{table}
\subsection{Main Results}

\paragraph{Results on Interactive Tasks} Table \ref{table1} demonstrates the strong performance of \textbf{tlm-DRE} on both ALFWorld and SciWorld. As shown, prompt-only ReAct baselines achieve only moderate results: GPT-4 attains an average reward of 38.1 in ALFWorld (unseen) and 64.4 in SciWorld (unseen), while GPT-3.5 lags significantly behind. Qwen2.5-7B performs modestly in all settings, with an average reward of around 25, indicating limited effectiveness without further alignment or training.

For SFT and RL training, most prior work adopts LLaMA2-7B as the backbone. To ensure a maximally fair comparison, we also report results with LLaMA2-7B. We further evaluate a stronger and more recent model, Qwen2.5-7B. ETO directly applies DPO to agent tasks, avoiding the need for a critic network as in PPO and thus offering a lighter and simpler training pipeline. With LLaMA2-7B, ETO achieves a score of 72.4 in ALFWorld (\textit{unseen}) and 61.1 in SciWorld (\textit{unseen}), although it is weaker in the seen setting. DMPO performs particularly well on SciWorld with LLaMA2-7B, reaching 72.4 on SciWorld (\textit{seen}), but degrades on ALFWorld. Based on the same base model, our method attains the best results in the seen-tasks on both datasets; moreover, when switching to Qwen2.5-7B, it further boosts ALFWorld (\textit{unseen}) to 90.1, far surpassing the LLaMA2-7B counterpart.

The main results presented in the paper focus primarily on off-policy methods. Nevertheless, given the recent surge of interest in on-policy approaches, such as GRPO \cite{shao2024deepseekmath}, we also compare our method with several state-of-the-art on-policy algorithms. The results are reported in Appendix~\ref{appendix_table}. 

We primarily select GiGPO \cite{feng2025group} as a representative on-policy method for comparison. However, GiGPO adopts a different evaluation protocol that involves longer exploration phases, which may lead to discrepancies in performance assessment. To ensure a fair comparison, we further evaluate our method under the same experimental setup and environment used by GiGPO. As shown in Appendix~\ref{appendix_table}, our approach outperforms GiGPO by approximately three points in terms of overall success reward, demonstrating its strong effectiveness even in this more demanding setting.   

\paragraph{Results on Multi-Hop QA Tasks} As shown in Table~\ref{tabel3}, \textbf{tlm-DRE} consistently improves performance on multi-turn search-augmented QA tasks, achieving an Exact Match (EM) of 33.8 and an F1 score of 43.74 in HotpotQA, substantially outperforming strong baselines such as NAT. Although search-augmented QA uses a different set of tools and typically requires fewer exploration steps than traditional agent tasks (e.g., ALFWorld), it implies generalization of our method.

\begin{figure}[htbp]
   \includegraphics[width=1.0\columnwidth]{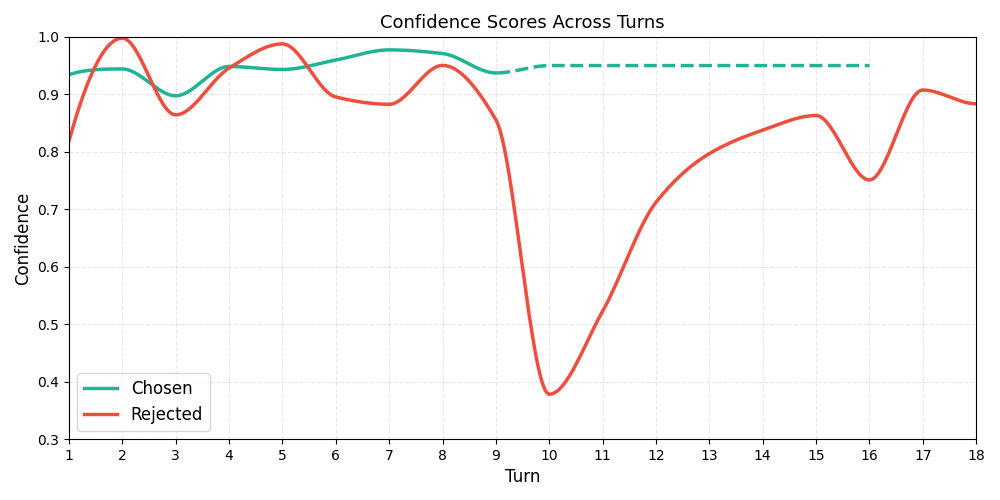}
   \caption{Illustration of confidence scores for different turns computed using the SFT policy, shown for both the chosen and rejected trajectories.}
   \label{fig:turn-confidence}
\end{figure}

\paragraph{Ablation Study} We conduct ablation studies to evaluate the effectiveness of each component: (1) Density Ratio Estimation (DRE) with UKL-kernel for Bregman divergence. (2) A variant version of DRE proposed with turn-level multiscale. Table~\ref{table1} shows that both components yield clear improvements over standard SFT and DPO. Specifically, turn-level weights and DRE each contribute independently, but exhibit different strengths across evaluation splits. We observe that the turn-level weights tend to provide larger gains in the \textit{unseen} setting on both datasets, suggesting that emphasizing critical turns helps to better fit out-distribution interaction patterns. Moreover, \textbf{tlm-DRE} brings more consistent improvements in \textit{unseen} tasks and achieves the best overall performance.

\section{Analysis}

\subsection{Illustration of turn weights}

Previous methods, such as ETO and DMPO, often view the entire trajectory as a unified object, thereby neglecting the contribution gaps of different turns within the trajectory. For example, in a failed trajectory, not every turn is necessarily incorrect. Therefore, as shown in Eq.~\eqref{Adequacy_Estimation}, we use the SFT policy to recompute the confidence scores of the model in different turns for both positive and negative examples. As illustrated in Fig.~\ref{fig:turn-confidence}, we present the confidence scores of the SFT policy on different turns for both chosen expert trajectories and rejected trajectories. We observe that, in the rejected trajectories, the model exhibits notably low confidence at Turns 10 and 11. By examining the chosen expert trajectories and specific case studies of Fig.~\ref{fig:case}, we find that these two turns are indeed critical steps that lead the entire trajectory to the wrong direction. Consequently, in the subsequent alignment stage, we place particular emphasis on such uncertain turns, and our experimental results confirm that this idea is effective.

\begin{figure}[htbp]
   \includegraphics[width=1.0\columnwidth]{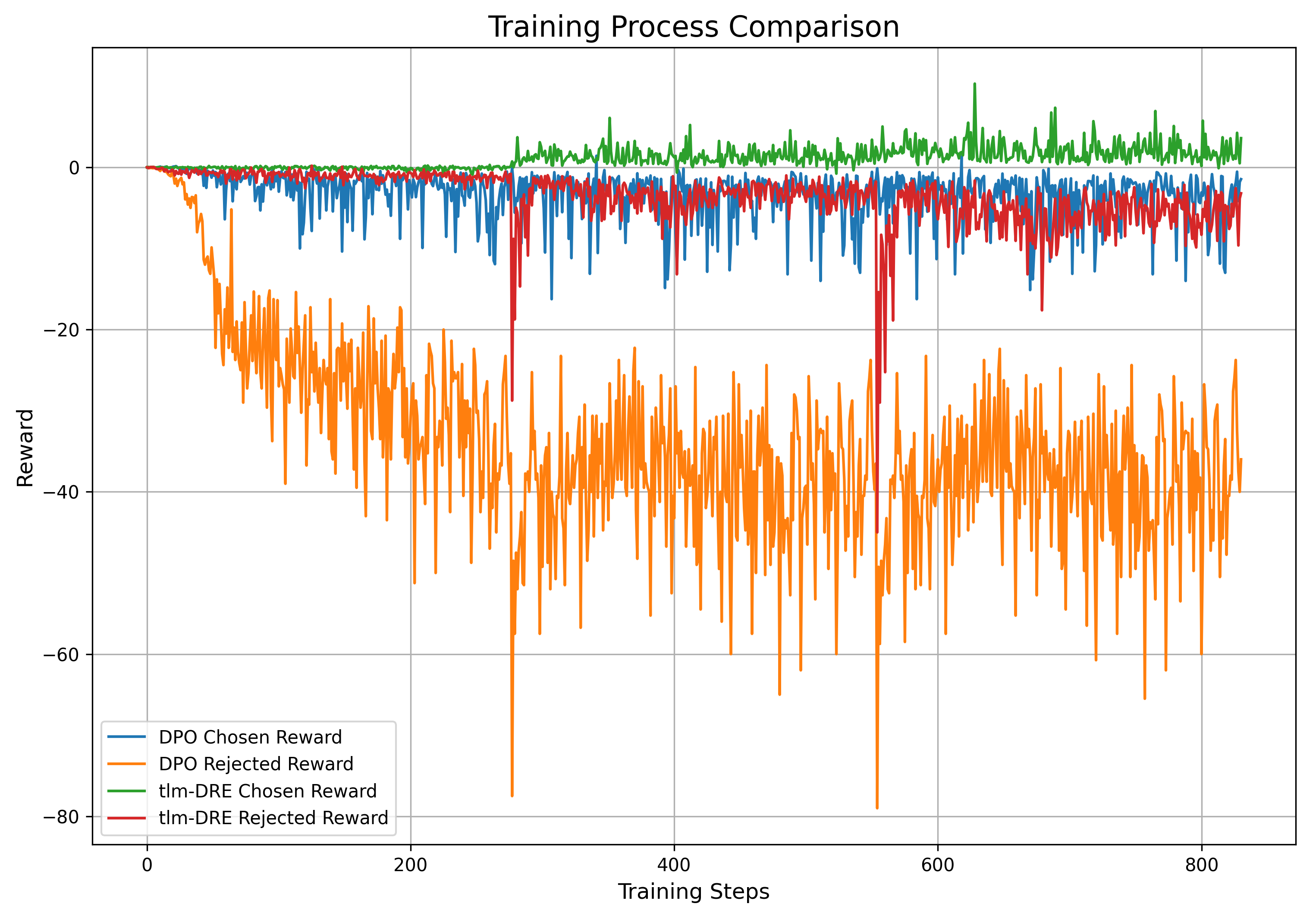}
   \caption{Illustration of the reward margin between the chosen and rejected trajectories during training for tlm-DRE and DPO}
   \label{fig:sf-exp3}
\end{figure}

\subsection{Margin analysis}

Fig.~\ref{fig:sf-exp3} shows the reward margin between the chosen and rejected trajectories during training for tlm-DRE and DPO. We argue that DPO separates chosen and rejected trajectories in a coarse-grained manner, treating a failed trajectory as entirely negative and neglecting that most turns within it can still be correct. As a result, many correct turns from rejected trajectories may be pushed toward the negative region, which can degrade overall performance.

To address this issue, our method up-weights critical turns and implicitly down-weights correct turns to be punished. Concretely, we only decrease the reward for the key erroneous turns while keeping the weights of other turns almost unchanged. As illustrated in Fig.~\ref{fig:sf-exp3}, our method produces a chosen-rejected margin smaller than the standard DPO, reflecting a finer-grained distinction between positive and negative samples. The empirical results further support the effectiveness of this design.

\section{Conlusions}
In this paper, Turn-level Multiscale Density Ratio Estimation (\textbf{tlm-DRE}) is proposed to enhance LLM performance in multi-turn agent tasks. We theoretically design a multiscale density ratio representation that leverages the contrastive information between positive and negative samples. Our approach assigns turn-specific weights under the framework of DRE to make the alignment more flexible and robust. Our approach has demonstrated strong performance across multiple multi-turn agent tasks. In the future, we will explore this method across more multi-turn scenarios.

\section{Limitations}
Our Turn-level Multiscale Density Ratio Estimation (\textbf{tlm-DRE}) employs pair-level asymmetric training with turn weights assigned to key steps, which adapts well to multi-turn agent tasks with long trajectories. However, this study is still limited from several perspectives, pointing to promising directions for future research. First, \textbf{tlm-DRE} is conducted mainly based on the offline sampling strategy, traditionally adopted by PPO or DPO. Whether the gain is still maintained when we combine the proposed alignment method with the dynamic sampling strategy used in GRPO or GSPO deserves careful investigation. Second, while the UKL kernel of Bregman divergence shows advances in several multi-turn agent tasks, systematic analysis of the impact of varying the choice of different function kernels needs to be further conducted. Moreover, with the rapid development of the AI/LLM agent area, there are more agent-based tasks emerging in recent years, which need more complex task decomposition and tool-using strategy, as well as more sophisticated environmental interaction. Extending the experiment of our method to more relevant tasks also warrants an in-depth analysis. We hope that our work will inspire researchers to explore multi-step agent training in this field.

\bibliography{custom}

\clearpage

\appendix

\section{Details of Density Ratio Estimation}
\label{appendix_kernal}

\begin{figure}[htbp]
   \includegraphics[width=1.0\columnwidth]{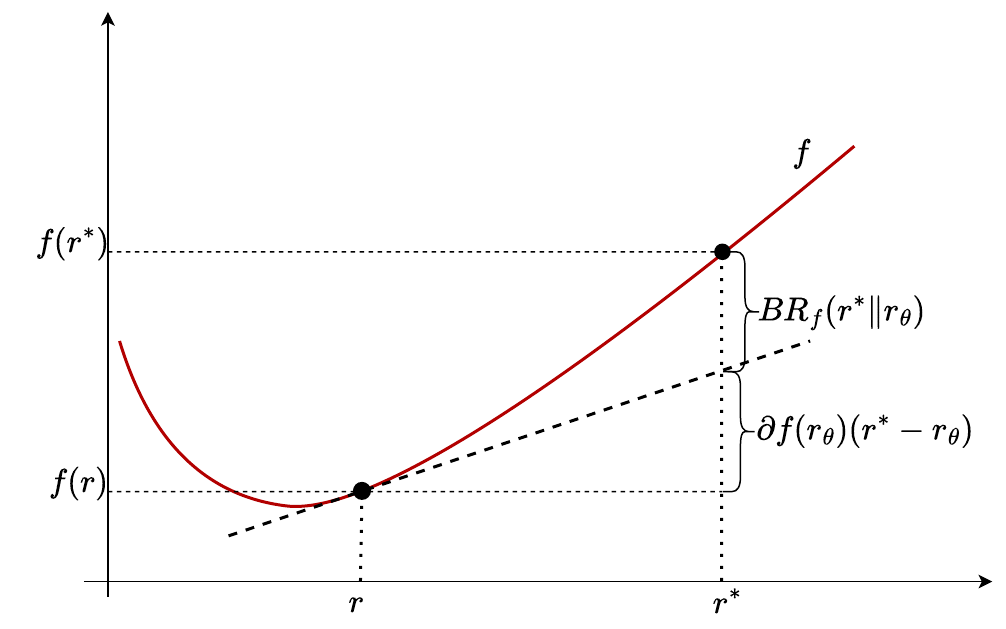}
   \caption{Illustration of DRE with Bregman divergence.}
   \label{fig:bregman_divergence}
\end{figure}

Fig.~\ref{fig:bregman_divergence} shows how optimizing the Bregman divergence gradually drives the point toward the target point.

Several kernel functions for Bregman divergence have been proposed in the past discourse, such as LSIF \citep{kanamori2009least}, UKL \citep{nguyen2010estimating}, BCE \citep{hastie2009elements}, and Basu’s power divergence \citep{basu1998robust}. We list the details of those functions below.

\paragraph{LSIF}
\begin{equation}
\begin{aligned}
f(r) = \frac{1}{2} (r - 1)^{2}.
\end{aligned}
\end{equation}
Bregman divergence (BR) defined in Equation~\ref{eq_breg_dre} is reduced to the squared distance(SQ):
\begin{equation} 
\begin{aligned} 
SQ^{\prime}(r^{*} \Vert r_{\theta})  = \frac{1}{2}(r^* - r)^2.
\end{aligned}
\end{equation}

\paragraph{UKL}
\begin{equation}
\begin{aligned}
f(r) = r\mathop{\mathrm{log}}r - r.
\end{aligned}
\end{equation}
BR is reduced to the unnormalized Kullback–Leibler (UKL) divergence:
\begin{equation} 
\begin{aligned} 
UKL^{\prime}(r^{*} \Vert r_{\theta})  = r^*\mathop{\mathrm{log}}\frac{r^*}{r} - r^* + r.
\end{aligned}
\end{equation}

\paragraph{BCE/BKL}
\begin{equation}
\begin{aligned}
f(r) = r\mathop{\mathrm{log}}r - (r + 1)\mathop{\mathrm{log}}(r+1).
\end{aligned}
\end{equation}
BR is reduced to the binary Kullback–Leibler (BKL) divergence:
\begin{equation} 
\begin{aligned} 
BKL^{\prime}(r^{*} \Vert r_{\theta})  = (1 + r^*)\mathop{\mathrm{log}}\frac{1+r}{1+r^*} + r^*\mathop{\mathrm{log}}\frac{r}{r^*}.
\end{aligned}
\end{equation}

\paragraph{Basu’s power} For $\alpha > 0$,
\begin{equation}
\begin{aligned}
f(r) = \frac{r^{1 + \alpha} - r}{\alpha}.
\end{aligned}
\end{equation}
Then BR is reduced to the BA divergence:
\begin{equation} 
\begin{aligned} 
BA_\alpha^{\prime}(r^{*} \Vert r_{\theta})  = r^\alpha(r - r^*) - \frac{r^*r^\alpha - (r^*)^{1 + \alpha}}{\alpha}.
\end{aligned}
\end{equation}

\section{Additional Experimental Results}
In addition to off-policy methods, we also compare against recent on-policy approaches. Our method still achieves competitive performance, demonstrating its effectiveness across different training paradigms.
\label{appendix_table}

\begin{table}[H]
\centering
\begin{tabular}{cc}
\hline
 & ALFWorld \\
\multirow{-2}{*}{Method} & (all) \\ \hline

RLOO \cite{ahmadian2024back} & 75.5 \\
GRPO \cite{shao2024deepseekmath} & 77.6 \\
GiGPO w/ std \cite{feng2025group} & 90.8 \\
GiGPO w/o std \cite{feng2025group} & 90.2 \\ \hline
tlm-DRE (ours) & \textbf{92.4} \\ \hline
\end{tabular}
\caption{For fair comparison, results are evaluated in the GiGPO \cite{feng2025group} test environment (with longer exploration steps and different test categories). All methods use the same base models: Qwen2.5-7B-Instruct}
\label{tabel2}
\end{table}

The main reason we did not include GiGPO's result Table \ref{table1} is that GiGPO's experimental setup is completely different from Table \ref{table1}. The main differences are as follows:
1. Different exploration steps during testing: Prior works typically adopt a maximum exploration horizon of 40 steps for ALFWorld and 10 steps for WebShop. In contrast, GiGPO extends this limit to 50 and 15 steps, respectively.
2. Different evaluation method: Taking ALFWorld as an example, prior works follow a standard protocol where the test set is partitioned into 140 seen (in-distribution) and 134 unseen (out-of-distribution) samples. In contrast, GiGPO employs a fundamentally different experimental setup. Instead of splitting by seen/unseen status, it curates a specific subset of 120 samples, categorized into six groups based on action types.
3. Different datasets were selected: it lacks the result of ScienceWorld, which is a very important multi-turn dataset put forward as an agentic benchmark from ReAct.

\section{Experiment Setup}
\label{appendix_setup}

We mainly select \texttt{Qwen2.5-7B-Instruct} \cite{team2024qwen2} for experiment. For a complete comparison, we also selected \texttt{Llama2-7B-Chat} \cite{touvron2023llama}. Our model is fully fine-tuned (not PEFT) in two stages: 3 epochs of SFT followed by 1 epoch of alignment training, optimizing with AdamW \cite{kingma2014adam}. For SFT, the initial learning rate is $1\times10^{-5}$ for  Alfworld, Sciworld, and $2\times10^{-5}$ for HotpotQA.  For alignment, the initial learning rate is $7\times10^{-7}$ for Alfworld and $1\times10^{-6}$ for Sciworld, HotpotQA.
In the alignment stage, policy sampling uses temperature 1 with a batch size of 4, while inference testing uses temperature 0. We launch alignment training 3 times for each task and compute the statistical information of the performance as shown in Table \ref{table1}.



\begin{figure}[htbp]
   \includegraphics[width=1.0\columnwidth]{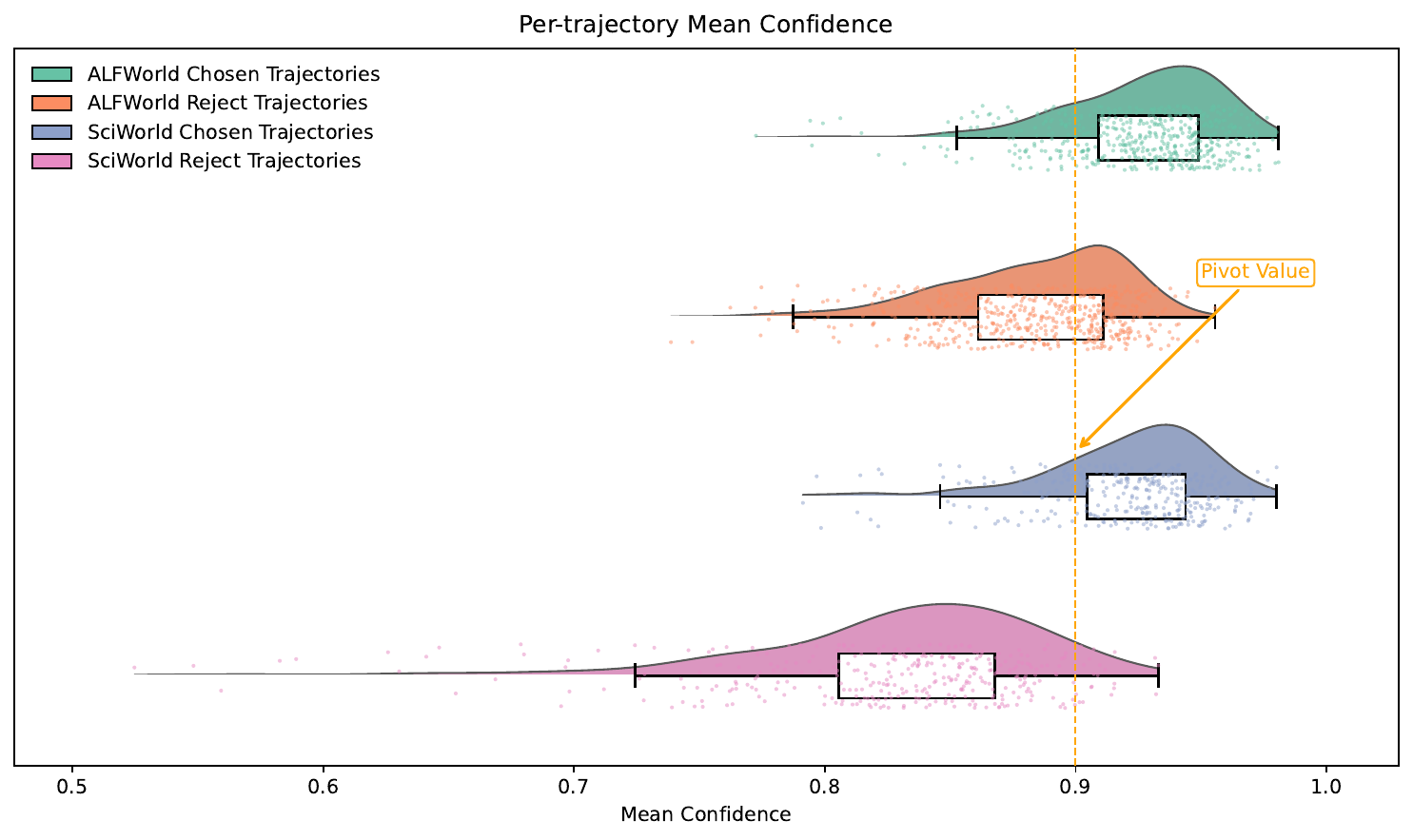}
   \caption{The statistical information of the adequacy/confidence score defined in Equation~\ref{Adequacy_Estimation} for each trajectory. The orange dashed vertical line indicates the selected pivot value $p_{pivot}$.}
   \label{fig:distribution_trajectory_mean_confidence}
\end{figure}

\begin{figure}[htbp]
   \includegraphics[width=1.0\columnwidth]{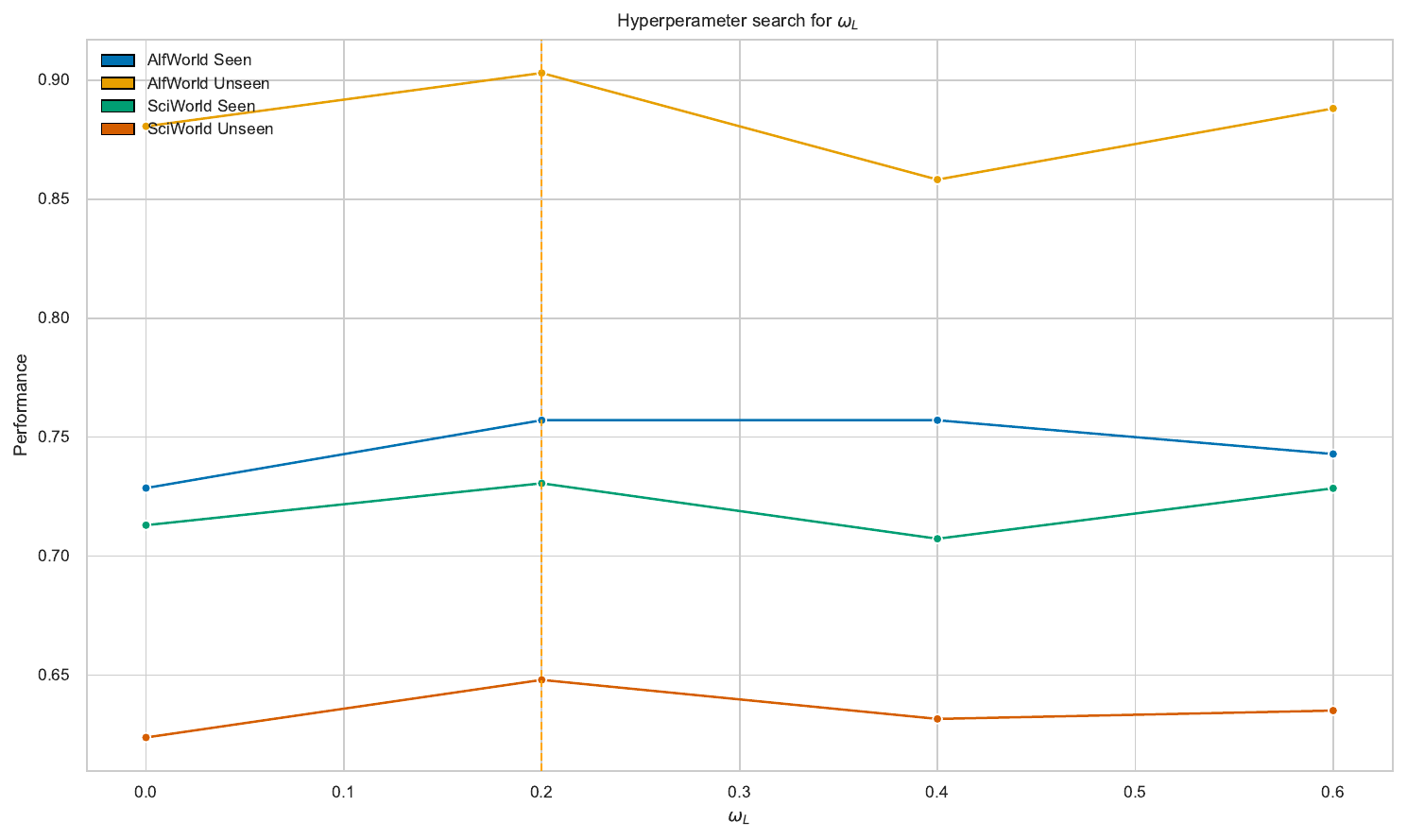}
   \caption{The searching result of the hyperparameters $\omega_L$ in the main agentic tasks. The dashed vertical line shows the optimal value of $\omega_L$}
   \label{fig:hyper_down_search}
\end{figure}

For the selection of hyperparameters $\omega_L$, $\omega_U$, $p_{pivot}$, we set the value of $p_{pivot}$ at 0.9 heuristically, as illustrated in Fig.~\ref{fig:distribution_trajectory_mean_confidence}. Most of the reference policy's generation probabilities are near this value except when a critical error turn occurs in the trajectory. For $\omega_U$, it is set as 1.0 based on the logic that when critical error turns occur, RL needs to train heavily on those turns so that the procedure of RL will downgrade back to the standard DPO or the standard DRE when $\omega_U$, is 1.0. On the other hand, $\omega_L$ plays as a low-temperature warming coefficient, making the other turns not be trained too much in the RL stage, while maintaining the relatively same reward margin as in the SFT stage between the chosen samples and the rejected samples(as shown in Fig.~\ref{fig:sf-exp3}). Hence, you are correct, we do some hyperparameter search only on , we set $\omega_L$ as 0.0, 0.2, 0.4, and 0.6 as shown in Fig.~\ref{fig:hyper_down_search}. We chose the best value of 0.2 for all testing datasets.

\section{Other Topics}
\label{appendix_other_topics}

\subsection{When Confidence-based Turn Weights Work}
\label{appendix_why_confidence}

In Figure~\ref{fig:distribution_trajectory_mean_confidence}, we show the statistical information of the adequacy/confidence score for both chosen and rejected samples. There is an apparent gap between the chosen and rejected samples, which is why we argue that lower confidence, especially after SFT, is more likely to indicate under-fitting instead of hallucination. And this under-fitting is, besides adding more data, more related to token efficiency during post-training. 

On the other hand, directly applying this idea to a raw policy's confidence score of the output tokens, which is not further tuned in the specific domain as is often the case for online policy RL, remains to be doubtfully working well. In such a scenario, even though the confidence score can be recomputed frequently based on the updated policy during the post RL training, whether the stubborn high-score hallucination tokens can be fixed by providing massive training dataset and high-frequent trajectories sampling remains an open question, as we discussed in the limitation section.

Therefore, it is recommended to start the analysis of the confidence distribution between the chosen/rejected samples, as shown in Figure~\ref{fig:distribution_trajectory_mean_confidence},  before the tlm-DRE post training.

\subsection{Comparison With Other RL/IL Methods}
\label{appendix_compare}

\paragraph{DIL} 
The prior DIL focuses on the general benchmark for LLM, such as MMLU, etc. Hence, it does not include any agentic tasks in its experiments. We reproduced DIL's results on agentic tasks: the "tlm-DRE w/o tlm" shown in Table~\ref{table1}. We also draw a conclusion based on our experimental results that the vanilla DRE(or named DIL) contributes more in unseen tasks while the turn-level weights tend to provide larger gains in the seen tasks on both datasets, suggesting that critical turns helps better fit in-distribution interaction patterns.

\paragraph{ReAct/Reflexion/Search-R1}
There are many other ReAct style templates. For example, Reflexion modifies the ReAct template by adding the rethinking steps, Search-R1 introduces the ReAct style template on the multi-turn QA tasks. Our work focuses on demonstrating that simulation learning with turn weight is a token efficiency post-alignment method for general kinds of multi-turn tasks, Therefore, we only apply the original ReAct format to make sure that the experiment is not only customized for a certain domain. Meanwhile, our method is potentially compatible with all these methods, which focus more on how to design the planning/thinking procedure extending from ReAct.

\paragraph{GRPO/GSPO/GiGPO}
Since our method is essentially more akin to imitation learning, it is more suitable for approximating an existing golden or chosen expert trajectories, which fits better into the offline RL paradigm. It is hard to pick one as the single expert in the dynamic sampling batch for GRPO in a certain step.  On the other hand, some experiment results have already shown that simply launching the vanilla GRPO is not a silver bullet for all tasks. For example, Table~\ref{tabel2} shows that our result surpasses the previous online-policy method in the ALFWorld tasks. Search-R1 shows the advantages of PPO over online policy RL in QA tasks based on some LLM backbone. Therefore, we need to study more both theoretically and experimentally to design a compatible way to realize this combination in the future.

\section{Trajectory Sample}
\label{appendix_cases}

\begin{figure}[htbp]
  \centering
  \includegraphics[width=0.95\linewidth]{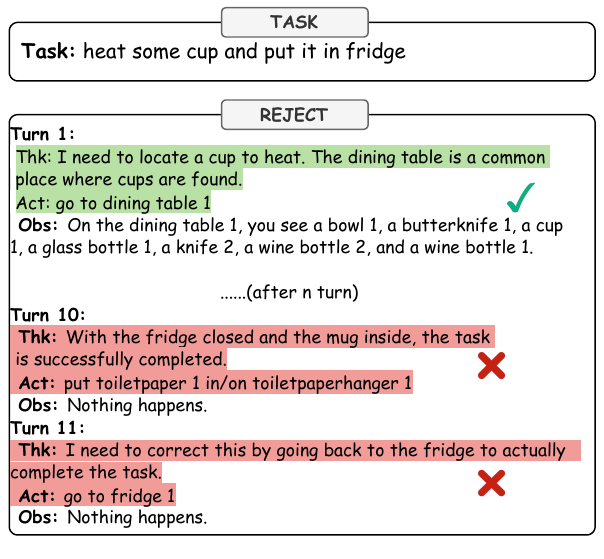}
  \caption{Case analysis across different turns.}
  \label{fig:case}
\end{figure}

In practice, the failure trajectory is often caused by only a few critical erroneous steps. For example, Fig.~\ref{fig:case} shows how different turns of a trajectory vary in their importance and completeness of training. For the rejected sample in Fig.~\ref{fig:case}, the first nine turns in this trajectory perform correctly, while the tenth turn produces an apparent wrong action, leading the following steps to deviate from the original goal.

\begin{figure}[htbp]
  \centering
  \includegraphics[width=0.95\linewidth]{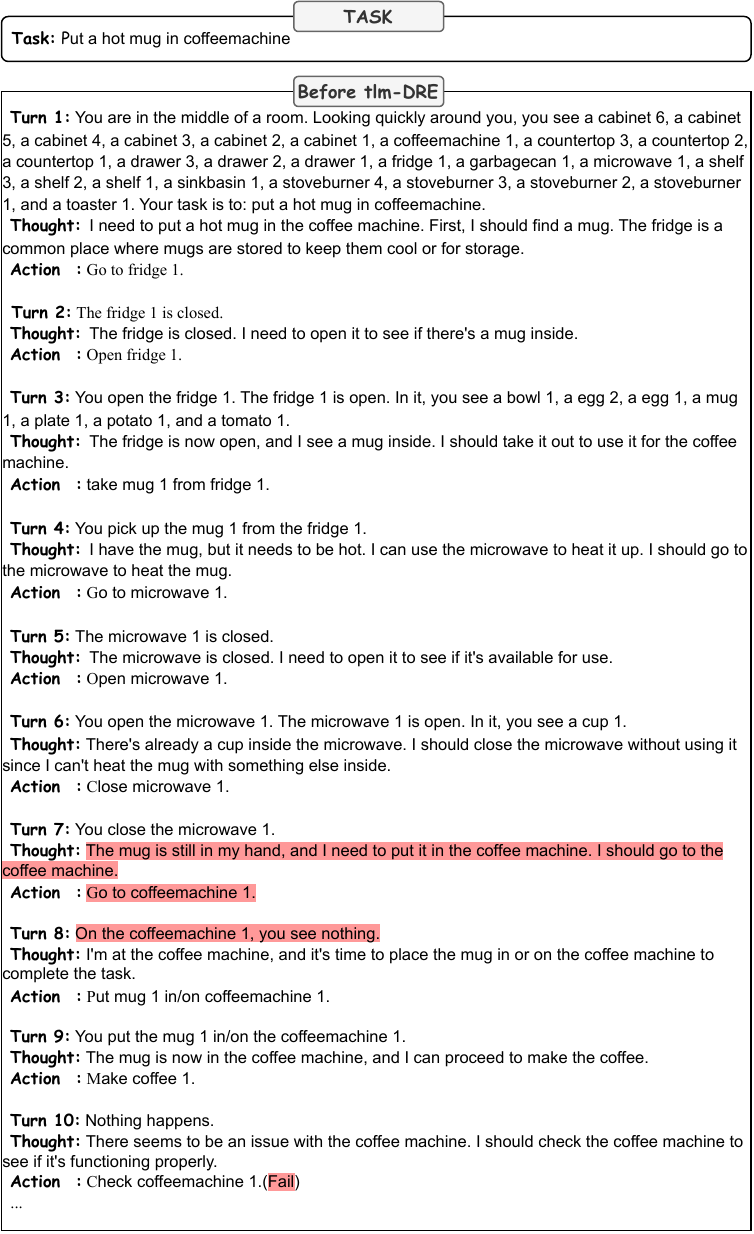}
  \caption{Trajectory sample before \textbf{tlm-DRE}.}
  \label{fig:case_before_align}
\end{figure}

\begin{figure}[htbp]
  \centering
  \includegraphics[width=0.95\linewidth]{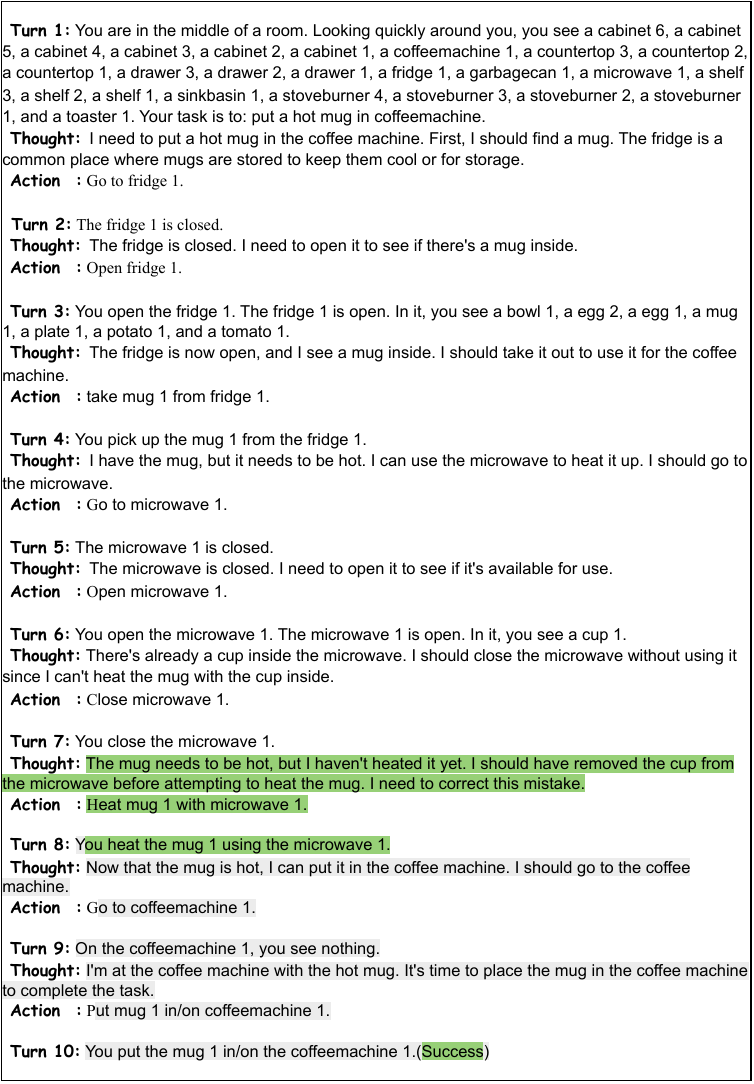}
  \caption{Trajectory sample after \textbf{tlm-DRE}.}
  \label{fig:case_after_align}
\end{figure}

Figs.~\ref{fig:case_before_align} and ~\ref{fig:case_after_align} also show how the key turn improved after alignment. In this case, the seventh turn plays an important role in the task, which is a crucial step that needs to be fixed. Therefore, the turns that truly need to be rectified are relatively sparse in the alignment stage. Hence, our proposed method is well-suited for agentic tasks that involve long turn trajectories, with higher token efficiency and less margin drifting, as shown in Fig.~\ref{fig:sf-exp3}.

\end{document}